\documentclass[runningheads]{utils/llncs}
\usepackage{utils/eccv}

\usepackage{graphicx}
\usepackage{amsmath,amssymb} 
\usepackage{utils/eccvabbrv}
\usepackage[accsupp]{axessibility}  
\usepackage{orcidlink}

\usepackage{amsmath}
\usepackage{amsfonts}
\usepackage{amssymb}
\usepackage{wrapfig}
\usepackage{subcaption}
\usepackage{multirow}
 \usepackage{mathtools} 

\usepackage{verbatim}

\usepackage{anyfontsize}

\usepackage{microtype}
\usepackage{graphicx}
\usepackage{booktabs} 
\usepackage{multirow}
\usepackage{amsmath,amssymb}
\usepackage{booktabs}
\usepackage{caption,subcaption}

\usepackage{xcolor}
\definecolor{mygreen}{HTML}{3cb44b}
\definecolor{skyblue}{HTML}{beffff}
\definecolor{lightgreen}{HTML}{90ee90}

\usepackage{color, colortbl}

\definecolor{emerald}{rgb}{0.31, 0.78, 0.37}

\usepackage{tcolorbox}
\usepackage{enumitem}
\setitemize{itemsep=10pt,topsep=0pt,parsep=0pt,partopsep=0pt}
\usepackage{colortbl}

\usepackage{xcolor}
\definecolor{mygreen}{HTML}{3cb44b}
\colorlet{myyellow}{green!10!orange!90!}
\makeatletter

\usepackage{tikz}
\usetikzlibrary{arrows,shapes,snakes,automata,backgrounds,fit,petri}
\usepackage{adjustbox}

\newcommand{\RN}[1]{%
	\textup{\lowercase\expandafter{\it \romannumeral#1}}%
}
\usepackage{tabu}

\newcommand{\beq}{\vspace{0mm}\begin{equation}}
\newcommand{\eeq}{\vspace{0mm}\end{equation}}
\newcommand{\beqs}{\vspace{0mm}\begin{eqnarray}}
\newcommand{\eeqs}{\vspace{0mm}\end{eqnarray}}
\newcommand{\barr}{\begin{array}}
\newcommand{\earr}{\end{array}}

\newcommand{\Xmat}[0]{{{\bf X}}}

\newcommand{\xv}{\boldsymbol{x}}

\newcommand{\thetav}{\boldsymbol{\theta}}

\usepackage{color, colortbl}
\definecolor{Gray}{gray}{0.93}

\usepackage{lipsum}

\newcommand\blfootnote[1]{%
  \begingroup
  \renewcommand\thefootnote{}\footnote{#1}%
  \addtocounter{footnote}{-1}%
  \endgroup
}

\usepackage{pifont}
\newcommand{\xmark}{\ding{55}}%

\usepackage{makecell}

\usepackage{xcolor,amsmath}
\usepackage[linesnumbered,ruled,vlined]{algorithm2e}
\DontPrintSemicolon

\usepackage{xcolor}
\definecolor{mygreen}{HTML}{3cb44b}

\SetKwComment{Comment}{\color{green!50!black}\# }{}

\SetKwProg{Function}{def}{:}{}

\SetKwProg{For}{for}{:}{}
\SetKwProg{If}{if}{:}{}

\definecolor{deeporange}{RGB}{255,140,0}
\newcommand{\mydeeporange}[1]{\textnormal{\ttfamily\color{deeporange}#1}\unskip}
\definecolor{deeppurple}{RGB}{128,0,128}
\newcommand{\mydeeppurple}[1]{\textnormal{\ttfamily\color{deeppurple}#1}\unskip}

\definecolor{lightgray}{gray}{0.9}

\usepackage{booktabs}
\usepackage{color}
\usepackage{multirow}
\definecolor{lightgreen}{rgb}{0.6, 0.9, 0.6}
\definecolor{lightred}{rgb}{0.9, 0.6, 0.6}

\usepackage{tikz,pgfplots}
\usepackage{marvosym}

\newcommand{\our}{\texttt{MovieSeq}}

\newcommand{\mad}{MAD}

\newcommand{\cmd}{CMD}
\newcommand{\movieqa}{MovieQA}
\newcommand{\movienet}{Movienet}

\newcommand{\tvc}{TVC}

\newcommand{\LLM}{Large Language Models}
\definecolor{darkgreen}{RGB}{0,128,0}
\definecolor{darkred}{RGB}{128,0,0}

\definecolor{NiceBlue}{rgb}{0.11764705882352941, 0.5647058823529412, 1.0}

\hypersetup{pagebackref,breaklinks,colorlinks,citecolor=eccvblue}

\begin{document}
\pagestyle{headings}
\mainmatter
\def\ECCV16SubNumber{***}  

\title{Learning Video Context as\\Interleaved Multimodal Sequences} 
\titlerunning{MovieSeq}

\author{Kevin Qinghong Lin$^1$,~
Pengchuan Zhang$^2$,~
Difei Gao$^1$,
Xide Xia$^2$,\\
Joya Chen$^1$,~
Ziteng Gao$^1$,~
Jinheng Xie$^1$,~
Xuhong Xiao$^3$,~
Mike Zheng Shou$^1$\textsuperscript{\Letter}
}
\authorrunning{Lin et al.}

\institute{
$^1$Show Lab, National University of Singapore\quad $^2$Meta AI\quad $^3$DSO National Laboratories\\
}

\maketitle

\vspace{-2em}
\begin{figure}[!h]
  \centering
  \includegraphics[width=1.0\textwidth]{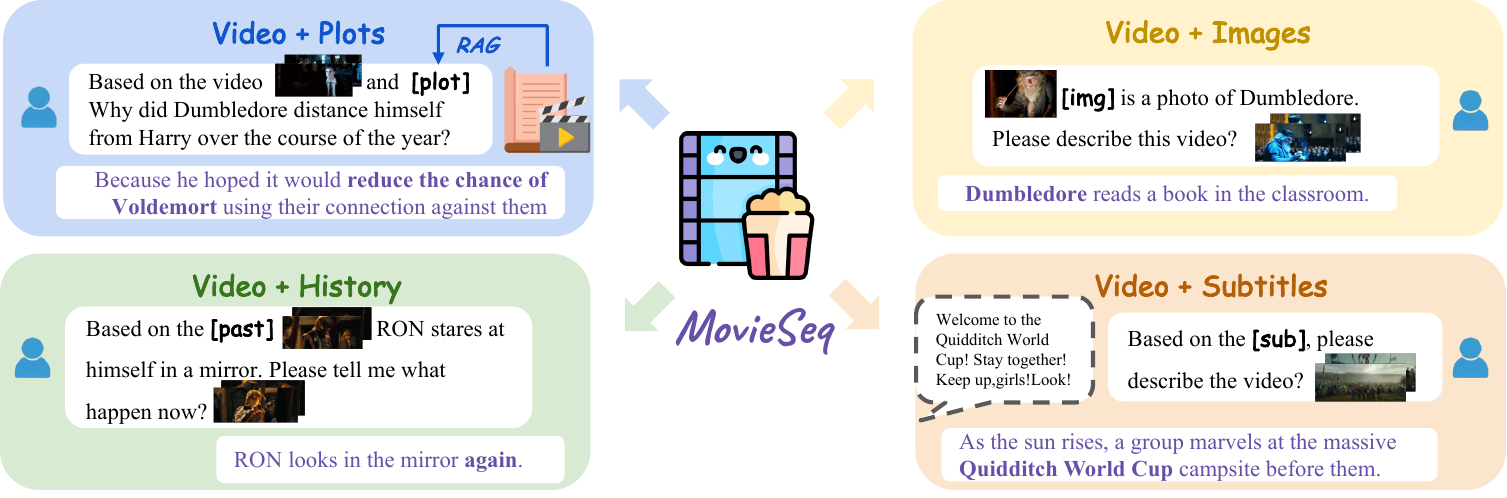} %
    \captionof{figure}{
    {\textbf{\our}} aims to address diverse challenges in understanding video contexts, enabling flexible interleaved multimodal instructions, such as Video+Images (for character identification), Video+Subtitles (for dialogues understanding), Video+Plots (for external knowledge via RAG), and Video+History (for event dependency).
    }
    \vspace{-2.5em}
  \label{fig:movieseq} %
\end{figure}

\begin{abstract}
Narrative videos, such as movies, pose significant challenges in video understanding due to their rich contexts (characters, dialogues, storylines) and diverse demands (identify who~\cite{autoad2}, relationship~\cite{lfvu}, and reason~\cite{movieqa}).
In this paper, we introduce~{\our}, a multimodal language model developed to address the wide range of challenges in understanding video contexts.
Our core idea is to represent videos as interleaved multimodal sequences (including images, plots, videos, and subtitles), either by linking external knowledge databases or using offline models (such as whisper for subtitles).
Through instruction-tuning, this approach empowers the language model to interact with videos using interleaved multimodal instructions. 
For example, instead of solely relying on video as input, we jointly provide character photos alongside their names and dialogues, allowing the model to associate these elements and generate more comprehensive responses.
To demonstrate its effectiveness, we validate \our's performance on six datasets (LVU, MAD, Movienet, CMD, TVC, MovieQA) across five settings (video classification, audio description, video-text retrieval, video captioning, and video question-answering). 
The code will be public at {\url{https://github.com/showlab/MovieSeq}}.

\end{abstract}
\blfootnote{\Letter: Corresponding Author.}

\section{Introduction}
Narrative videos, like movie clips, TV series, etc., offer a lens into our diverse world, depicting human stories across history and culture through extended visual streams. These videos pose unique challenges such as 
character identification~\cite{lsmdc,movienet,autoad2,storyboard}~(e.g., recognizing ‘Who’), 
situation understanding~\cite{movieqa,cmd,lfvu,moviegraphs}~(e.g., relationship and dialogue), and event dependency~\cite{mad,autoad,egoschema}~(e.g., cause-and-effect over time). Yet, most top-performing video perception models~\cite{slowfast,videomae,internvideo} typically prioritize on concise videos, emphasizing atomic objects~\cite{davis,objectron,ytbbox} and actions~\cite{hmdb,kinetics,something}.

Significant progress in understanding video contexts has been made via varied setups such as video recognition~\cite{activitynet,lfvu}, video-text retrieval~\cite{lsmdc,cmd}, and video question-answering~\cite{movieqa,egoschema}.
These advances include:
efficient transformer architecture to support extended video durations, applied in tasks like classification~\cite{longterm,memvit,timesformer} and video question-answering~\cite{mist,zsvqa};
enhanced temporal event association for video-text retrieval~\cite{milnce,egovlp,univtg}, localization~\cite{tan,univtg} and video-captioning~\cite{mart,videollmol};
audio modality integration for enhanced situational understanding~\cite{eclipse,umt,merlot}; and the
use of external knowledge, as exemplified by \cite{autoad2} in training a character recognition module by actors's databases~\cite{movienet} for audio descriptions.
Despite these advancements, prior methods remain task-specific, with design limitations persisting primarily due to:
(i) The diversity of contexts (e.g., images, videos, texts, subtitles) and tasks (captioning, retrieval, question-answering, etc).
(ii) The tailored technical designs required to adapt to particular settings such as \cite{autoad2} train extra modules to access external knowledge.
Having witnessed the advancements in \LLM~(LLM)~\cite{chatgpt,gpt4,llama,llama2}, the question arises: 
can we develop a general solution that handles these diverse contexts and needs in videos?

Although LLMs exhibit versatility in natural language processing~\cite{llama,llama2,gpt4} and multi-modal scenario~\cite{openflamingo,llava,minigpt}, developing a LLM for complex video understanding is not straightforward.
Pioneering efforts such as~\cite{blip2,llava,minigpt,videochat} project single visual input (e.g., image or video) to textual tokens space in conjunction with user queries, as shown in Fig. \ref{fig:relatedwork}a.
Meanwhile, studies like~\cite{flamingo,otter,cosmo} (see Fig. \ref{fig:relatedwork}b) introduce interleaved in-context learning to improve the model's few-shot capabilities by providing structured demonstration. 
However, when applied to narrative videos, which encompass informative contexts 
, these models with a pre-defined visual-textual template still exhibit limitations due to inflexibility. This highlights the need for a more flexible approach to handling multimodal contexts within videos.

Motivated by these observations, we develop {\our}, a multimodal language model that is designed for narrative video understanding. Acknowledging the diverse contexts and tasks in videos, our core concept is to embed the videos as interleaved multi-modal sequences. 
As depicted in Fig. \ref{fig:movieseq} and Fig. \ref{fig:relatedwork}c, our approach unifies various multimodal contexts (images, subtitles, plots, video history) and tasks into a user-friendly sequence, subsequently processed by the language model.
Moreover, one of the key challenges is the lack of instruction-following data for complex videos. We present a packaged solution on how to convert existing video datasets for interleaved multimodal instruction-following. Finally, we present a decoder-only multi-modal language model, which is trained on our constructed data, to support a variety of task types.

To demonstrate the effectiveness of \our, we conducted experiments across six video benchmarks (LVU~\cite{lfvu}, MAD~\cite{autoad}, \movienet~\cite{movienet}, CMD~\cite{cmd}, TVC~\cite{tvqa}, MovieQA~\cite{movieqa}) across various settings (video classification, audio description, character identification, video-text retrieval, video captioning, and video question-answering).
Additionally, \our~facilitates a novel application, allowing users to interact with video using free-form interleaved multi-modal instructions.
Overall, our contributions are three folds:

\begin{figure}[!t]
  \centering
  \includegraphics[width=1.0\textwidth]{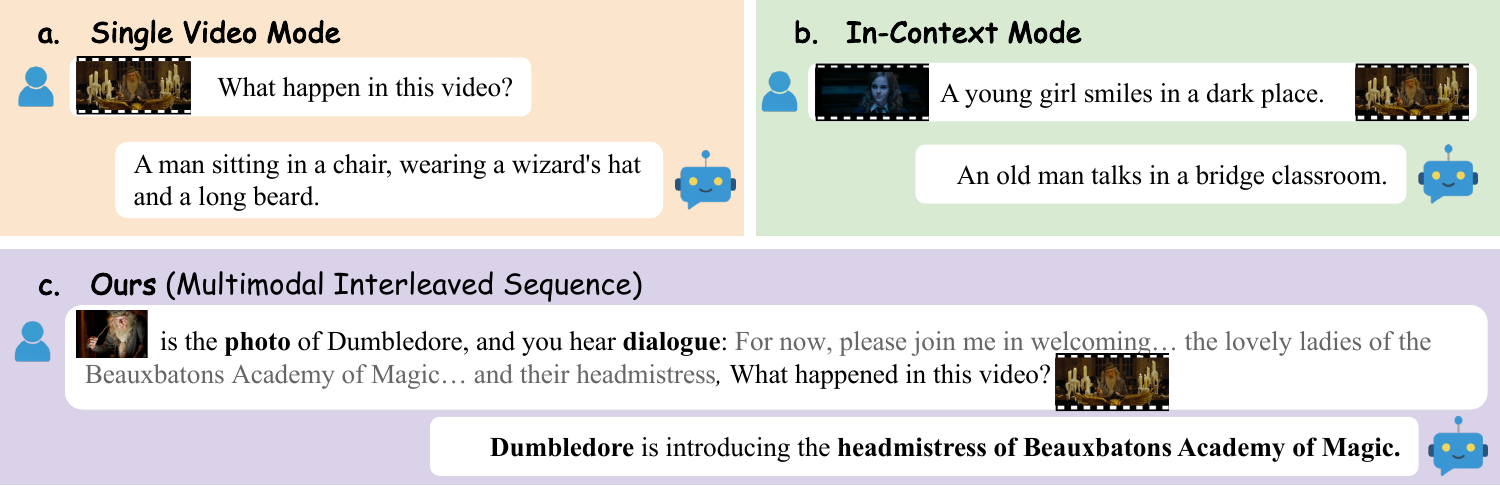} %
  \caption{\textbf{Comparison between different video-language input modes.} 
(a) Single video input, e.g.,\cite{llava,minigpt,videochat}.
(b) In-context input, e.g.,\cite{flamingo,otter}, showcasing examples for structured few-shot learning.
(c) Our approach, utilizes flexible contexts (e.g., external character images, dialogues, etc) to associate them to produce a comprehensive response.
\vspace{-3em}
} %
  \label{fig:relatedwork} %
\end{figure}

\begin{enumerate}[itemsep=0.pt, topsep=-1pt, leftmargin=12pt]    
    \item We propose \our, a video-language model that embeds videos into interleaved multimodal sequences to flexibly adapt to diverse contexts. 
    \item We present a package of solutions for converting videos into various forms of interleaved multimodal instructions (e.g., video-images, video-plots, etc).
    \item We demonstrate the flexibility and effectiveness of \our, one generative model, which performs well across five tasks and six datasets.
\end{enumerate}
\vspace{-0.4cm}
\section{Related Work}
\vspace{-0.2cm}

\noindent
\textbf{Movie Understanding} seeks to build an interpretation of narrative video streams, which goes beyond basic object and action recognition. This presents challenges in achieving high-level video understanding and reasoning.
In the visual unimodal domain, various works~\cite{vis4mer,moviesst,s5} have focused on improving network architecture for efficient long-form video recognition.
When comes to the multimodal domain, there has been a range of benchmarks focusing on advanced aspects of video contexts, encompassing topics like characters~\cite{lsmdc,mad}, relationships~\cite{moviegraphs,lfvu}, emotions~\cite{emotion}, dialogues~\cite{cmd,tvqa}, storylines~\cite{movienet,vidchapters}.
To address these challenges, substantial efforts have been made.
Studies like 
\cite{cliphit,vila,moviept} primarily enhance video-text alignment along the temporal axis. 
\cite{mart} employs a memory module to improve video captioning's coherency.
\cite{eclipse} introduces sound integration to assist video retrieval.
Recent AutoAD-II~\cite{autoad2} trains a character recognition module using external image databases~\cite{movienet} for precise automatic AD generation.
Notwithstanding this progress, most methods predominantly focus on individual tasks with tailored architectures, illustrating the complexity of video contexts and tasks.
Our work seeks one model solution to support various modalities contexts, unifying diverse tasks in a generative manner.

\noindent
\textbf{Large Multimodal Models.}
The advent of LLMs~\cite{gpt2,gpt3,gpt4} has significantly impacted natural language processing and inspired numerous studies~\cite{assistgpt,mmreact,vipergpt} in multimodal domains.
Several studies~\cite{blip2,llava,minigpt,instructblip,videochat} have pioneered Large Multimodal Models by projecting the single visual input (e.g., an image or video) into textual embedded spaces, subsequently aligning them with the LLMs (e.g., Llama~\cite{llama,llama2}) by visual instruction tuning.
Additionally, various efforts such as \cite{flamingo,otter,cosmo}, have investigated interleaved vision-text input modes, aiming to augment language models with in-context, few-shot demonstrations.
More recently, several studies~\cite{shikra,gpt4roi} have focused on enhancing visual spatial understanding e.g., models are capable of producing responses with coordinates.
A number of work~\cite{moviechat,llamavid} have proposed efficient architectural solutions to address the memory challenges posed by long video durations.

Nevertheless, the complexity of narrative videos, which include diverse elements such as characters photos, dialogue, and external metadata, continues to pose challenges. Recent work by \cite{mmvid,mmnarrator} has proposed the use of GPT-4V~\cite{lmm_cv} to transform video streams into textual document format. 
In our work, we intend to utilize an open-source language model, e.g., Llama2~\cite{llama2} with interleaved multimodal instruction tuning for various video applications.
\section{Challenges in Video Contexts}
\label{sec:longvideo}
\noindent
In this section, we illustrate challenges  and needs associated with understanding contexts in narrative videos.
We choose movies as a representative example due to their descriptive storylines, informative metadata such as characters, and extended durations.

\noindent
\textbf{(i) Situational Dialogue.} 
While visuals serve as a basic medium for conveying information, achieving a comprehensive understanding of the context demands that the model accommodates audio inputs such as dialogues. This requires the models to associate and interpret the significance of dialogue in conjunction with the visual stream.
For instance, when provided with a video of a grassland featuring a girl and the dialogue ‘Finally, I arrive at grandmother's ranch.’ the model should generate an informative narration like ‘A girl arrives at her grandmother's ranch.’
Therefore, this necessitates the model to have the ability to concurrently decode dialogue and establish visual associations.

\noindent
\textbf{(ii) Event Dependency.}
Recognizing event dependencies is crucial for understanding video storylines, as these often contain fine-grained events with causal links between earlier and later events. For instance, in a sequence of movie clips~\cite{autoad}, if an old man retrieves a flashlight from a car, it indicates his intention to illuminate something later. 
This type of dependency helps viewers construct a narrative bridge that connects past, present, and future elements in the video.
Therefore, it is essential for the video model to manage not only individual events but also to link multiple separate events, implying the need to jointly accept one or more video clip inputs and establish associations.

\noindent
\textbf{(iii) External Knowledge.} 
Leveraging external information is often beneficial and necessary to improve the comprehension of videos, e.g., before watching a movie, viewers often find it helpful to access concise metadata like the movie's title, genre, plot summary, or glimpses of characters through trailers. It requires the assistant to flexibly integrate various external knowledge, e.g., characters' photos (visual) or movie plot documents (textual).

The above challenges are not exclusive to movies; they are widely applicable to other videos e.g., those social media videos usually involve additional contextual elements such as video titles, descriptions, user avatars, and ASR transcripts.

\begin{figure*}[t]
  \centering
  \includegraphics[width=1.0\textwidth]{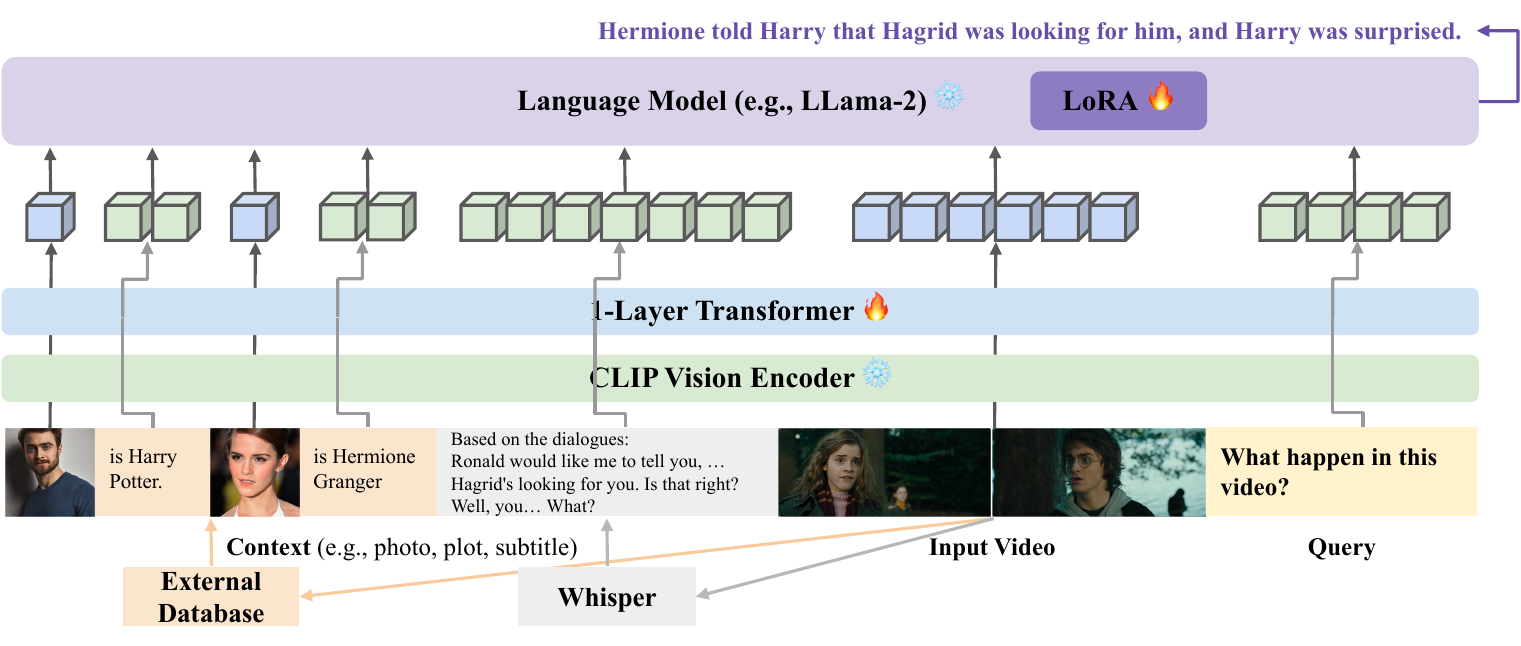} 
  \caption{
  Illustration of the pipeline of \our.
  Firstly, we embed the input video as an interleaved multimodal sequence (such as images, plots, videos, or subtitles), either by linking to an external database or leveraging annotations from offline models~\cite{whisper,whisperx}.
  Then, we create an interleaved instruction (can be a combination of the above context) and feed it into the language model. The language model is trained to associate them and generate a comprehensive response.
  }
  \label{fig:pipeline} 
\vspace{-2em}
\end{figure*}

\vspace{-0.4cm}
\section{Methodology}
\vspace{-0.2cm}

We first illustrate \our~pipeline and architecture, then provide more details of interleaved multimodal instruction construction.

\vspace{-0.4cm}
\subsection{Overview}
\noindent
\textbf{Equip Video with Contexts.}
As shown in Fig.\ref{fig:pipeline}(bottom), we first transform input videos as an interleaved multimodal sequence with additional context.
For external knowledge, this includes character photos sourced from public datasets\cite{autoad2,movienet} or textual movie plots from IMDb.
For dialogues, we obtain speech transcriptions from offline models such as whisper~\cite{whisper,whisperx}, ensuring temporal alignment with the video.

\noindent
\textbf{Architecture.}
We illustrate our architecture in Fig.~\ref{fig:pipeline}, building upon the common framework outlined in~\cite{llava}.
We utilize the pre-trained LLama-2~\cite{llama2} as our language model, integrating LoRA~\cite{lora} into all linear layers to achieve efficient and effective tuning. We freeze all parameters except those in the input embedding layers, resulting in a mere 0.12\% increase in trainable parameters by LoRA.
The visual encoder is CLIP~\cite{clip} ViT-B/16.
To handle multiple images and videos of varying lengths, we introduce two modifications:
(i) For visual inputs, we extract frame-wise \texttt{[CLS]} token instead of patch tokens.
(ii) We project the visual embeddings through a single-layer transformer with temporal position encoding to capture temporal relationships.
We recognize a performance constraint by not adopting patch tokens as in previous works~\cite{videochat,videochatgpt}. While our focus is on exploring benefits in additional contexts. This enables efficient handling of multiple videos of varying lengths and image inputs, by treating images as a single frame, thereby enabling the flexibility of handling free-form interleaved inputs.

\vspace{-1em}
\subsection{Interleaved Multi-Modal Instructions}
Our primary focus is on leveraging available contexts to create an interleaved multi-modal instruction for narrative videos. To this end, we first introduce the generic instruction format, as detailed in Tab.~\ref{tab:input_sequence}, which comprises three terms:
(i) context $\Xmat_{\texttt{ctx}}$, which facilitates multi-modal interleaved context, encompassing elements like characters photos, dialogues, storylines, or video history;
(ii) question $\Xmat_{\texttt{q}}$, represents the user query, e.g., ‘Can you help me describe this video?’ or as a formulation of the downstream task, e.g., ‘What is the relationship in this video?’
(iii) answer $\Xmat_{\texttt{a}}$ denotes the model's responses.
This design allows the same question $\Xmat_{\texttt{q}}$ to yield varied responses $\Xmat_{\texttt{a}}$ under different contexts $\Xmat_{\texttt{ctx}}$.

\begin{table}[h!]\centering
\begin{minipage}{0.99\columnwidth}
\vspace{-2em}    
\centering
\begin{tcolorbox} 
    \raggedright
    \small
    $\texttt{USER}:$ \mydeeporange{$\Xmat_{\texttt{ctx} }$}
    $\Xmat_{\texttt{q}}$$\textbackslash \texttt{n}$ $\texttt{MovieSeq}$: 
    \mydeeppurple{$\Xmat_{\texttt{a} }$} 
    \mydeeppurple{\texttt{<EOS>}} \\
\end{tcolorbox}
    
\caption{
\textbf{Our interleaved multimodal instruction} comprises the context \mydeeporange{$\Xmat_{\texttt{ctx}}$}, the question $\Xmat_{\texttt{q}}$, and the answer \mydeeppurple{$\Xmat_{\texttt{a}}$}. The context term provides necessary multimodal context (e.g., an arrangement of image, text, video, subtitles), and the model is trained to predict the answer \mydeeppurple{$\Xmat_{\texttt{a}}$} followed by \mydeeppurple{\texttt{<EOS>}} to indicate termination.
\vspace{-2em}
}
\label{tab:input_sequence}
\end{minipage}
\end{table}

\vspace{-1em}

\noindent
\textbf{Video with Images.}
Considering the frequent needs for character identification~\cite{autoad,tvqa,lsmdc} in movies, we exemplify our video with image instruction through this task. 
Practically, we identified two demands for character identification.
Firstly, we need to determine which characters, whether one or several, appear in the video;
Secondly, we want to interact with the video, supported by the identified character photos.
Based on these requirements, we have devised the following two contexts and questions\footnote{We use grey color to represent the context e.g., \colorbox{lightgray}{\texttt{img}}: image tokens; \colorbox{lightgray}{\texttt{vid}}: video tokens; \colorbox{lightgray}{\texttt{sub}}: subtitles; \colorbox{lightgray}{\texttt{plot}}: selected movie plot.}:

\noindent
\fbox{\parbox{\dimexpr\linewidth-2\fboxsep-2\fboxrule\relax}{
{(i$_a$)} $\Xmat_{\texttt{ctx}}:$ There are several character photos: 
\colorbox{lightgray}{\texttt{img$_1$}} is \colorbox{lightgray}{\texttt{name$_1$}}, \colorbox{lightgray}{\texttt{img$_2$}} is \colorbox{lightgray}{\texttt{name$_2$}}, ..., \colorbox{lightgray}{\texttt{img$_n$}} is \colorbox{lightgray}{\texttt{name$_n$}} and a video \colorbox{lightgray}{\texttt{vid}}. \\
$\Xmat_{\texttt{q}}:$ Who can be found in this video? If not, output None.\\
{(i$_b$)} $\Xmat_{\texttt{ctx}}:$  There have character photos: \colorbox{lightgray}{\texttt{img$_1$}} is \colorbox{lightgray}{\texttt{name$_1$}} (can have more) and a video \colorbox{lightgray}{\texttt{vid}}.
\\
$\Xmat_{\texttt{q}}$: Please briefly describe this video (or other free-form query).}}

With these two designs, we can identify who is in the video as well as generate a response with character identification.

\noindent
\textbf{Video with Plots.}
Movies come with a wealth of textual metadata, such as plots~\cite{movieqa} or synopsis~\cite{movienet}, which are valuable references for high-level comprehension. 
We chose MAD dataset~\cite{mad,autoad}, and gathered movie plot synopsis from imdb website~\footnote{https://www.imdb.com/}.
 to fulfill our needs.
However, the plot resembling a document tends to be quite long, and may surpass the context length of LLMs, accurately identifying relevant sections is crucial. We provide two solutions for plot sampling:
\noindent
\textbf{(a)} For instructions with a specific query such as a detailed question $\Xmat_{\texttt{q}}$ in MovieQA~\cite{movieqa}. We adopt retrieval-augmented approaches~\cite{ral,rag}. We obtain sentence embeddings\footnote{We use \texttt{all-mpnet-base-v2} from SentenceTransformers.}, perform sentence-paragraph retrieval using cosine similarities, and subsequently return the top-1 relevant paragraph as the context $\Xmat_{\texttt{ctx}}$. 

\noindent
\textbf{(b)} For general queries such as clip captioning, we process each video clip centered at timestamp $t$ in a movie with duration $T$ by computing its ratio $r=\lfloor \frac{t}{T} \rfloor$. This ratio $r$ is then utilized to pinpoint the $r$-th sentence in the plot and extract a paragraph using a sentence number as the window size $w$.

With the sampling plots, we define the instructions as:

\noindent
\fbox{\parbox{\dimexpr\linewidth-2\fboxsep-2\fboxrule\relax}{
{(ii)} $\Xmat_{\texttt{ctx}}:$ Based on the plot \colorbox{lightgray}{\texttt{plot}} and the video \colorbox{lightgray}{\texttt{vid}}.
}}
where the $\Xmat_{\texttt{q}}$ can be any free-form questions.

\noindent
\textbf{Video with Subtitles.}
Subtitles serve as a unique signal, offering extensive situation information, including human emotions, intentions, and specific terminology, and etc.
We chose CMD~\cite{cmd} and TVC~\cite{tvqa} as data sources, as each video clip in their datasets is paired with temporal aligned subtitles and captioning. This captioning, such as ‘Beckett talks to Montgomery about her date and looks over at her coworkers’, effectively links visuals with subtitles, making them suitable for supervising instruction-tuning. We design such a context:

\noindent
\fbox{\parbox{\dimexpr\linewidth-2\fboxsep-2\fboxrule\relax}{
{(iii)} $\Xmat_{\texttt{ctx}}:$ Based on the \colorbox{lightgray}{\texttt{sub}} and the video \colorbox{lightgray}{\texttt{vid}}.
}}

\noindent
\textbf{Video with History.}
In long narrative videos, capturing temporal dependencies i.e., associations among multiple events is crucial. Leveraging historical videos as context, can enhance the understanding of the current or aid in predicting future events.
For example, if a man is seen purchasing a train ticket, this suggests he may later appear at a station, thereby adding coherence to the narrative.
MAD~\cite{mad} is known for its long continuous narrations, which we employ to formulate the following instructions: 

\noindent
\fbox{\parbox{\dimexpr\linewidth-2\fboxsep-2\fboxrule\relax}{
{(iv)}
$\Xmat_{\texttt{ctx}}:$ There are $n+1$ video clips, ordered from the past to present: \colorbox{lightgray}{\texttt{vid$_\text{t-n}$}} \colorbox{lightgray}{\texttt{cap$_\text{t-n}$}}
$\cdots$
\colorbox{lightgray}{\texttt{vid$_\text{t-2}$}} \colorbox{lightgray}{\texttt{cap$_\text{t-2}$}} 
\colorbox{lightgray}{\texttt{vid$_\text{t-1}$}} \colorbox{lightgray}{\texttt{cap$_\text{t-1}$}} 
\colorbox{lightgray}{\texttt{vid$_\text{t}$}}.
}}

In the above instruction, we include both the history video and their narrations \colorbox{lightgray}{\texttt{vid$_\text{t-i}$}} \colorbox{lightgray}{\texttt{cap$_\text{t-i}$}} as context together, which we find beneficial. 
The past narration can be derived from either annotations or the model's previous predictions in a recurrent inference setting. 

Notably, the above four instruction templates are designed for videos with one context. For videos with multiple contexts, such as MAD~\cite{mad,autoad} and TVC~\cite{tvqa}, we expand the instructions by concatenating these contexts. 
Additionally, we utilize ChatGPT-3.5 to rephrase templates, aiming to achieve diversity and improve robustness. For example, the query: ‘who can be found in this video?’ with answer: \texttt{name} is transformed into a boolean question: ‘Does \texttt{name} appear in this video?’ with answer: ‘Yes’.

\vspace{-0.4cm}
\subsection{Training Objectives}
\vspace{-0.1cm}

\noindent
Despite the interleaved multimodal instruction designs, the training loss is computed based on \mydeeppurple{$\Xmat_{\texttt{a}}$}.
We employ the common language modeling training objective, i.e., auto-regressive, as follows:
\begin{equation}
    \max\sum_{i=1}^{L}\log p( {\color{deeppurple}\Xmat_{\texttt{a}}} | \Xmat_{\texttt{ctx}}, \Xmat_{\texttt{q}})=
    \prod_{i=1}^{L} p_{\thetav} (  {\color{deeppurple} \xv_i}
    |\Xmat_{\texttt{ctx}}, \Xmat_{\texttt{q}}, {\color{deeppurple} \Xmat_{\texttt{a}, <i}}
    ),
\label{eq:auto_regressive}
\end{equation}
\noindent
where $L$ the length of model response sequence i.e., \mydeeppurple{$|\Xmat_{\texttt{a}}|$}.
\vspace{-0.4cm}
\section{Experiments}
\vspace{-0.2cm}

\begin{table}[t]
\centering
\scriptsize
{%
\begin{tabular}{@{}p{0.15\linewidth}p{0.1\linewidth}p{0.15\linewidth}p{0.2\linewidth}p{0.15\linewidth}p{0.2\linewidth}@{}}
\toprule
Datasets & \#Sample & Video Len. & Task    & Metric &  Used Contexts  \\ \midrule
LVU~\cite{lfvu} &  1.5K  & 154.0s & Video Classification & Acc.  & vid, {sub} \\
\movienet~\cite{movienet} & 77K & key frames & Multi-label Cls. & F1-score &  vid, img\\
MADv2~\cite{autoad} & 300K & 4.1s-1.8h* & Audio Description & R-L, C, etc & vid, {img}, {plot}, hist.\\
\cmd~\cite{cmd} & 24K & 132.0s & Video Retrieval & Recall@N  & vid, {sub}\\
\tvc~\cite{tvqa} & 174K & 9.1s & Video Captioning & B4, R-L, etc. & vid, {sub}, {img}\\
MovieQA~\cite{movieqa} & 9.8K & 202.7s & VideoQA & Acc. & vid, sub, {plot}\\
\bottomrule
\end{tabular}
}
\caption{
\textbf{Dataset statistics.} Datasets are used for creating instruction and evaluation. 
These datasets vary in duration (ranging from keyframes to several seconds to minutes) and encompass diverse tasks with diverse contexts.
The notation * indicates that each movie (avg 1.8h) comprises multiple short clips (avg. 4.1s), so that we can leverage history context (hist).
}
\label{tab:datasets}
\end{table}

In this section, we structure our experiments to investigate the following questions:

\noindent{$\mathbf{Q1}.$ \textbf{Why LLM?}} Without additional data, can \our~exhibits flexibility and effectiveness compared with baselins across various setups?

\noindent{$\mathbf{Q2}.$ \textbf{Why Context?}} What impact does each video context have?

\noindent{$\mathbf{Q3}.$ \textbf{Why Instruction-tuning?}} Can we get a general model by multi-task multi-context instruction-tuning?

\noindent{$\mathbf{Q4}.$ \textbf{Is Data Construction Trivial?}} We perform ablation studies to examine setups in instruction construction, such as template design, sampling strategy.

\vspace{-0.4cm}
\subsection{Datasets and Settings}
\vspace{-0.2cm}
We assess \our~on six datasets spanning various tasks, as summarized in Tab.~\ref{tab:datasets}. For each setting, we briefly introduce the dataset and evaluation metrics. 
For non-generative tasks e.g., video classification (i,ii) and video-text retrieval (iv), we  express the adaptation of our generative language model to these predictions, with further details in the Supp.

\noindent
\textbf{(i) LVU}\cite{lfvu} comprises 30K videos from 3K movies, focusing on content understanding (relationship, speaking style, scene), metadata prediction (e.g., director, writer, year), and user engagement (e.g., Youtube like ratio). As \our~is a generative model, we concentrate on content understanding, which has a clear, inferable interpretation.
We employ Whisper~\cite{whisper} to extract subtitles for LVU videos. 
The metric employed is the top-1 accuracy. 
Our \our~addresses this classification task by directly generating the category names.

\noindent
\textbf{(ii) \movienet}~\cite{movienet} is a movie dataset that offers keyframes and character mappings for each frame. We adhere to the settings described in~\cite{autoad2} to examine the character identification, a multi-label classification task. 
Our \our~addresses this classification task by directly generating character names without the need for score thresholds. 
Thus, we opt for Recall, Precision, and F1-score as our evaluation metrics.

\noindent
\textbf{(iii) MADv2}~\cite{autoad} comprises a collection of 264K movie audio descriptions. Its goal is to automatically  generating narrations that align with individual video elements, incorporate character names, and ensuring temporal  coherence.
Following the official settings~\cite{autoad,autoad2}, we evaluate predictions against groundtruth using ROUGE-L~\cite{rouge} (R-L), CIDEr~\cite{cider} (C), SPICE~\cite{spice} (S), and Recall@k within N
neighbours (R@k/N)~\cite{autoad2} based on BertScores~\cite{bertscore}.
Notably, the above AD metric focuses on single captions and overlooks correlations among multiple sentences. As suggested by \cite{mart}, we employ Repetition@4 (Rep@4) to evaluate sentence repetition in our ablation studies.

\noindent
\textbf{(iv) \cmd}~\cite{cmd}, a long-range video-text retrieval benchmark, includes key scenes from over 3K movies, each paired by a high-level description, such as the movie's storyline. We align with the recent baseline~\cite{vila}, and adopt the Geometric Mean, R@1, R@5, and R@10 as metrics.
Although video-text retrieval is a contrastive task, we transform it by generating captions for each video. Then, we conduct sentence retrieval using these generated captions against ground-truth queries.

\noindent
\textbf{(v) \tvc}~\cite{tvqa} is a television captioning dataset, consisting of 262K descriptions matched with 108K video segments. TVC captions not only describe visual elements of the content but also incorporate dialogue elements from the subtitles. 
We follow \cite{swinbert} and adopt BLEU4~\cite{bleu} (B4), METEOR~\cite{meteor} (M), ROUGE-L~\cite{rouge} (R-L), and CIDER~\cite{cider} (C) as mertics.

\noindent
\textbf{(vi) \movieqa}~\cite{movieqa}, a video question-answering dataset, presents a challenge for the model to answer reasoned questions.
Each question offers five multiple-choice options, with accuracy serving as the evaluation metric.

\noindent
\textbf{Implementation Details.}
For MAD, we sample 8 frames per clip.
For \movienet, we sample each available frame.
For CMD, TVC, and MovieQA, we sample 32 frames per video.
For LVU, we sample 64 frames per video.
We train our model for 20 epochs on each dataset, using a learning rate of 3e-5. 
For LoRA~\cite{lora} settings, we use a rank of 16 with an alpha of 16.
For subtitles or plots, the maximum length is set to 512. 
For model responses, the default maximum length is set to 64. 

\subsection{Main Results (\textbf{Q1\&Q2}).}
In this section, we compare \our~on each benchmark individually, showcasing its flexibility and effectiveness relative to baseline methods. \textit{Notably, for a fair comparison, we report results using individual training set}. We will study the effect of multi-task instruction-training in ablation section.

\begin{table}[!t]
\centering
\scriptsize
{%
\begin{tabular}{@{}p{0.25\linewidth}p{0.12\linewidth}|p{0.14\linewidth}p{0.14\linewidth}p{0.14\linewidth}p{0.14\linewidth}@{}}
Methods      & Style &Relation  & Speak & Scene & \textbf{Avg.} \\ \midrule
VideoBERT~\cite{lfvu} & Discr. & 52.8 & 37.9 & 54.9 & 48.5\\
Obj. Tran.~\cite{lfvu} & Discr. & 53.1 & 39.4 & 56.9 & 49.8\\
VIS4mer~\cite{vis4mer} & Discr. & 57.1 & 40.8 & 67.4 & 56.9\\
LF-VILA~\cite{lfvu} & Discr. & 61.5 & 41.3 & 68.0 & 60.9\\
S5~\cite{s5} & Discr. & 67.1 & 42.1 & 73.5 & 67.0\\
MA-LMM~\cite{malmm} & Gen. & 58.2 & 44.8 & 80.3 & 61.1 \\
\midrule
\our~(vid) & Gen.  & 61.0 & 37.6 & \textbf{76.8} & 58.5\\
\our~(sub) & Gen. & 61.0 & 61.0 & 62.4 & 61.5  \\
\our~(vid, sub) & Gen. & \textbf{75.6}\textcolor{darkgreen}{$_{\uparrow14.6}$} & \textbf{63.0}\textcolor{darkgreen}{$_{\uparrow25.4}$} & 65.9\textcolor{darkred}{$_{\downarrow10.9}$} & \textbf{68.1}\textcolor{darkgreen}{$_{\uparrow9.5}$} \\
\end{tabular}
}
\caption{
\textbf{Video Classification} on LVU~\cite{lfvu} content understanding. Discr. means discriminative model e.g., classifier. Gen. means generative language model. 
The color \textcolor{darkgreen}{green} and \textcolor{darkred}{red} denote the gain or drop by the introduction of subtitles compared with the variant (vid).
\vspace{-3.5em}
}
\label{tab:main_lvu}
\end{table}

\noindent
\textbf{LVU.}
In Tab.~\ref{tab:main_lvu}, we initially evaluate \our, on the LVU content understanding test set. This is primarily a video classification task and no baseline has attempted to introduce subtitles. 
We observe that the  dialogue significantly aids in assessing relations (+14.6\%) and speech (+25.4\%), while it diminishes the scene understanding (-10.9\%).
This makes sense as dialogue usually reflects the situational content but does not directly benefit the purely visual aspects.

\begin{table}[h]
\centering
\scriptsize
\vspace{-2.5em}
{%
\begin{tabular}{@{}p{0.2\linewidth}p{0.2\linewidth}|p{0.15\linewidth}p{0.15\linewidth}p{0.15\linewidth}p{0.15\linewidth}@{}}
Methods     & Style &     Precision           & Recall & F1-score  \\ \midrule
CLIP cos.~\cite{clip} & Discr.  &  39.6  & 71.8 & 51.0  \\
TFM Decoder~\cite{autoad2} & Discr. &  75.9 & \textbf{82.7} & 79.1 \\
\our & Gen. &  \textbf{88.5} & 75.5 & \textbf{81.4} \\
\end{tabular}
}
\caption{
\textbf{Actor Identification} on \movienet~\cite{movienet}.
}
\vspace{-3.5em}
\label{tab:main_movienet}
\end{table}

\noindent
\textbf{\movienet.}
In Tab.~\ref{tab:main_movienet}, we evaluate character identification on \movienet, a multi-label classification task. 
The baseline results are sourced from ~\cite{autoad2}, where we select their score threshold by yielding the highest F1-score. It's worth noting that this task is a multi-label classification task.
Our method surpasses the baseline TDM Decoder~\cite{autoad2}, a module specifically tailored for this task, by 2.3 in F1-score.
With Tab.~\ref{tab:main_lvu} and Tab.~\ref{tab:main_movienet}, we demonstrate that the generative language model can perform well as a discriminative model. 

\begin{table}[b]
\vspace{-1.5em}
\centering
\scriptsize
\
{%
\begin{tabular}{@{}p{0.25\linewidth}p{0.2\linewidth}|p{0.1\linewidth}p{0.1\linewidth}p{0.1\linewidth}p{0.1\linewidth}p{0.1\linewidth}@{}}
Methods                         & Pretraining Data     & R-L & C   & S   & R@5/16 \\ \midrule
ClipCap~\cite{clipcap}  & CC3M   & 8.5 & 4.4 & 1.1 & 36.5  \\
CapDec*~\cite{capdec}     & AV-AD  & 8.2 & 6.7 & 1.4 & - \\
\midrule
AutoAD-I~\cite{autoad}  & -            & {{9.3}} & {6.7}          & {2.4}  & {-} \\
AutoAD-I~\cite{autoad}  & AV-AD \& WebVid  & {11.9}          & {{14.3}} & {{4.4}}           & {42.1} \\ 
\midrule
AutoAD-II~\cite{autoad2}  & -  & {12.7}          & {18.3} & {-}           & {45.6} \\ 
AutoAD-II~\cite{autoad2}  & AV-AD \& WebVid  & {13.4}          & {{19.5}} & {{-}}           & {50.8} \\ 
\midrule
\our~(vid,img,plot,hist.)&  -  & \textbf{15.5} & \textbf{24.4} & \textbf{7.0} & \textbf{51.6} \\
\end{tabular}
}
\caption{
\textbf{Audio Description generation} task on MAD-v2. 
The baseline results are from AutoAD-I\&II~\cite{autoad,autoad2}.
Please see Tab.\ref{tab:ablation_mad} for clear ablation of each context.
\label{tab:main_mad}
}
\end{table}

\noindent
\textbf{\mad.}
In Tab.~\ref{tab:main_mad}, we display the results on the MADv2 audio description benchmark.
Notably, this task requires identifying character names for each clip. Thus, we reuse the model trained on Tab.~\ref{tab:main_movienet} to predict character names. 
Our model, \our, without additional pretraining, achieves competitive results over the best baseline, i.e., a $4.9\%$ higher CIDEr.

\noindent
\textbf{\cmd.}
We carry out experiments on the \cmd~video-text retrieval task in Tab.~\ref{tab:main_cmd}.
In this task, we unexpectedly find that \our (vid) variant performs significantly poorly. 
Upon analyzing their predictions, we find that: \cmd~text queries such as `Max flies the drone home before his dad notices it's missing.' focus on the key scene storyline, including names, relationships, and different focal points.
The video-only generative tuning fails to find relevant cues to derive reasonable captions, resulting in a lot of hallucinations.
However, with the aid of dialogues, we significantly boost the performance, achieving a 29.74 Geo. Mean. improvement. 
This demonstrates that for a generative model, the quality of the tuning data is important (i.e., the answer should be derivable from the context).

\begin{table}[!t]
\centering
\scriptsize
{%
\begin{tabular}{@{}p{0.25\linewidth}p{0.15\linewidth}p{0.1\linewidth}|p{0.12\linewidth}p{0.1\linewidth}p{0.1\linewidth}p{0.1\linewidth}p{0.1\linewidth}@{}}
Methods     & \#PT Data & \#Frame & Geo. Mean           & R@1 & R@5  & R@10 \\ \midrule
MoEE~\cite{moee} & - & - & 5.9  & 1.9 & 7.8 & 13.4  \\
TeachText~\cite{teachtext} &-& -& 23.2  & 12.1 & 27.4 & 37.5  \\
Frozen~\cite{webvid} & 5M&32 & - & 12.6 & 28.4 & 36.3 \\
LF-VILA~\cite{vila} & 8M&32& 26.4 & 13.6 & 32.5 & 41.8 \\
VINDLU~\cite{vindlu} & 25M &32 & - & 18.4 & 36.4 & 44.3 \\
TESTA~\cite{testa} & 5M & 32 &- & 21.5 & 42.4 & \textbf{50.7} \\
\midrule
\our~(vid) & - & 32 & 9.2 & 4.3 & 11.8 & 14.9 \\
\our~(sub) & - & 32 & 33.1 & 21.2 & 38.0 & 44.8  \\
\our~(vid, sub) & - & 32 &  \textbf{38.9}\textcolor{darkgreen}{$_{\uparrow29.7}$} & \textbf{25.8}\textcolor{darkgreen}{$_{\uparrow21.5}$} & \textbf{45.3}\textcolor{darkgreen}{$_{\uparrow33.5}$} & {50.3}\textcolor{darkgreen}{$_{\uparrow35.4}$}  \\
\end{tabular}
}
\caption{
\textbf{Video-Text Retrieval} on \cmd~\cite{cmd} test set. 
The color \textcolor{darkgreen}{green} denotes the gain by the subtitles context compared with the variant (vid).
}
\vspace{-4em}
\label{tab:main_cmd}
\end{table}

\begin{table}[b]
\centering
\scriptsize
{%
\vspace{-2em}
\begin{tabular}{@{}p{0.2\linewidth}p{0.15\linewidth}p{0.1\linewidth}p{0.1\linewidth}|p{0.1\linewidth}p{0.1\linewidth}p{0.1\linewidth}p{0.1\linewidth}@{}}
Methods & Context & \#PT & \#Frame & B4 & R-L & M & C \\ \midrule
MMT~\cite{tvqa} & vid, sub & - & 3 FPS & 10.8 & 32.8 & 16.9 & 45.3 \\
HERO~\cite{hero} & vid, sub & 7.6M & 2/3 FPS & 12.3 & 34.1 & 17.6 & 49.9 \\
\textsc{Value}~\cite{value} &  vid, sub & - & 2/3 FPS & 11.6 & 33.9 & 17.6 & 50.5 \\
\textsc{SwinBERT}~\cite{swinbert} & vid & - & 64 & 14.5 & 36.1 & 18.5 & 55.4 \\
GIT$_\text{B}$~\cite{aio} & vid & 14M & 6 & 13.0 & 33.2 & 16.6 & 47.3 \\
All-in-one~\cite{aio} & vid & 105M & 3 & 12.5 & - & \textbf{20.4} & 56.3 \\
\bottomrule
\our & vid & - & 32 & 13.7 & 34.0 & 17.4 & 50.9 \\ 
\our & sub & - & 32 & 11.3 & 31.0 & 14.8 & 42.7 \\ 
\our & vid, sub  & - & 32 & {17.5} & {37.5} & 19.4 & {63.5} \\ 
\our & vid, sub, img$\dagger$ & - & 32 & \textbf{17.9}\textcolor{darkgreen}{$_{\uparrow4.2}$} & \textbf{38.1}\textcolor{darkgreen}{$_{\uparrow4.1}$} & 19.9\textcolor{darkgreen}{$_{\uparrow2.5}$} & \textbf{64.8}\textcolor{darkgreen}{$_{\uparrow13.9}$} \\
\end{tabular}
}
\caption{
\textbf{Video Captioning} on \tvc~\cite{tvqa} val set. Img$\dagger$ means the character's photos we collected. The color \textcolor{darkgreen}{green} denotes the gain by the subtitles and character photos compared with the variant (vid).
}
\label{tab:main_tvc}
\end{table}

\noindent
\textbf{\tvc.}
In Tab.~\ref{tab:main_tvc}, we validate \our~on the video captioning task, \tvc. 
This benchmark originally provides subtitles as inputs.
Without pretraining, \our~outperforms several baselines, including those utilizing subtitles such as HERO\cite{hero} and \textsc{Value}\cite{value}. This improvement demonstrates that language models with reasoning provide better visual-dialogue association.
Additionally, we found that the inclusion of character images does not significantly enhance performance. This could be because, in \tvc, the subtitles already provide the names (e.g., ‘Chandler: Yeah, well...’), and \tvc~only contains six TV series, thus having a limited number of characters, most are seen during training.

\newlength{\oldintextsep}
\setlength{\oldintextsep}{\intextsep}
\setlength{\intextsep}{5pt plus 2pt minus 2pt}


\begin{wraptable}{rh}{0.35\textwidth}

\scriptsize
{%
\begin{tabular}{@{}l|l@{}}
Methods     &     Acc. \\ \midrule
SSCB~\cite{movieqa}~(vid, subs) &  34.20  \\
PAMN~\cite{pamn}~(vid, subs) &  43.34  \\
HMMN~\cite{hmmn}~(vid, subs) &  46.28  \\
SSCB~\cite{movieqa}~(plot) &  56.70  \\
\midrule
\our~(vid) & 27.32 \\
\our~(subs) & 27.10 \\
\our~(plot) & 63.54 \\
\our~(vid, subs) & 48.32\textcolor{darkgreen}{$_{\uparrow21.9}$} \\
\our~(vid, plot) & \textbf{66.41}\textcolor{darkgreen}{$_{\uparrow39.1}$} \\
\end{tabular}
}
\vspace{-0.3cm}
\caption{
\textbf{Video QA} on MovieQA~\cite{movieqa} val set.
}
\label{tab:main_movieqa}
\end{wraptable}

\noindent
\textbf{MovieQA.}
In Tab.~\ref{tab:main_movieqa}, we validate \our~on the video question-answering task, \movieqa. This is a challenging setting, as the questions often contain types like ‘How’ and ‘Why’, which require reasoning beyond video. Thus, we observe that visual-only baselines yield weak performance (only 27.32 Acc.). 
We notice the gains of plots as they offer distilled information.
But our goal is not to exceed baseline scores via plot usage, but to offer new way for understanding video context within existing LLM methods, and show how to identify valuable parts from document-level plots.
This will be studied in ablation Fig.\ref{fig:context_len}(b).

\vspace{-1em}
\subsection{Ablation Studies}
\noindent
\textbf{Effect of Multimodal Contexts (Q2).}
In most previous tables, we have presented variants to ablate the impact of different modality contexts.
In Tab.~\ref{tab:ablation_mad}, we study the effect of different contexts (images, plot, history) for the audio description task.
There are several findings:
\textbf{(i)} From row 2, the external plot as a reference greatly improves the performance (+4.7 CIDEr).
\textbf{(ii)} From rows 3-4, with the past videos (with their narrations) as context, the model can effectively avoid repetition. If the past context from prediction (recurrent), it does not influence the original AD metrics. However, if the past context is sourced from the ground-truth (oracle), it can greatly boost them. This suggests that a reliable past context is more helpful.
\textbf{(iii)} Rows 5 and 6 in the table showcase distinct approaches: row 5 employs CLIP vision scores for character retrieval, whereas row 6 is the model trained with MovieNet~\cite{movienet}. This results indicate that character identification from additional instructions is necessary.
Lastly, with the above designs combined, we can jointly boost the AD metrics as well as avoid repetition. 

\begin{table}[!h]
\centering
\scriptsize
{%
\begin{tabular}{l|ll|llll|l}
\multirow{2}{*}{Rows} & \multirow{2}{*}{Settings} & \multirow{2}{*}{Context} & \multicolumn{4}{c|}{AD metrics} & \multicolumn{1}{l}{Multi-sentences} \\
 & & & R-L & C & S & R@5/16 & Rep@4 ~($\downarrow$)\\
\midrule
1 & Video & vid & 10.6 & 11.4 & 2.5 & 48.6  & 1.33 \\
2 & w/ Plot & vid, plot & 12.9 & 16.9\textcolor{darkgreen}{$_{\uparrow5.5}$} & 4.4 & 48.8  & 0.48\textcolor{darkgreen}{$_{\downarrow0.85}$} \\
3 & w/ Past & vid, hist. & 10.4 & 11.8\textcolor{darkgreen}{$_{\uparrow1.5}$} & 2.6 & 50.1  &  0.11\textcolor{darkgreen}{$_{\downarrow10.4}$} \\
\midrule
5 & w/ Characters (clip) & vid, img &  13.1 & 19.7  & 5.9 & 47.3 &  1.30 \\
6 & w/ Characters (movienet) & vid, img & 15.2 & 23.6\textcolor{darkgreen}{$_{\uparrow12.2}$}  & \textbf{7.2} & 50.3  & 0.97\textcolor{darkgreen}{$_{\downarrow0.36}$} \\
\midrule
7 & \our  & vid, img, plot, hist. & \textbf{15.5} & \textbf{24.4}\textcolor{darkgreen}{$_{\uparrow13.0}$}  & {7.0}  & \textbf{51.6}   & 0.12\textcolor{darkgreen}{$_{\downarrow12.1}$}  \\
\end{tabular}
}
\caption{\textbf{Ablative Studies of different context on \mad}. The color \textcolor{darkgreen}{green} denotes the gain by the subtitles and character photos compared with the variant (vid). For brevity, we only highlight the (C)IDER and Rep@4.
}
\label{tab:ablation_mad}
\end{table}

\noindent
\textbf{Effect of Multi-Task Co-Training (Q3).}
In Tab.~\ref{tab:cotraining}, we investigate the effect of multi-task co-training for a more robust model. 
Due to notable data imbalances across the datasets such as MAD being {200x} larger than LVU. 
We constructed a development set by selecting 1K samples from each dataset, including MAD (audio desc.), MovieNet (vid. cls.), and TVC (vid. cap.), to ensure comparable sizes with different tasks.

Multi-task co-training results in an average 1.5 improvement across three tasks, demonstrating that multi-task learning enhances individual capabilities.
This highlights the language model's ability to acquire commonsense across diverse objectives and contexts.

\begin{table}[!h]
\centering
\scriptsize
\setlength{\tabcolsep}{10pt}
{%
\begin{tabular}{l|c|l|l|l|l}
\multirow{2}{*}{{Training strategy}}  &  \multirow{1}{*}{{Multi}} &  {MAD} & {MovieNet} & {TVC}  & \textbf{Avg} \\
& {-task?} & (CIDEr) & (F1-score) & (CIDEr) &   \\
\hline
individual  & \xmark & 22.5 & 78.6 & 60.2 & 53.8 \\ 
co-training  & \ding{51} & {23.8} & {79.2} & {62.7} & {55.2}\textcolor{darkgreen}{$_{\uparrow1.5}$} \\
\end{tabular}
}
\caption{\textbf{Effect of multi-task co-training.}
These three datasets are similar in scale but encompass different tasks.
}
\vspace{-1.5em}
\label{tab:cotraining}
\end{table}

\noindent
\textbf{Strategy of Data Construction (Q4).}
In ‘Video with History’ setting, the template design and the history clip number play an important factor. Ablation studies conducted on the MAD benchmarks with oracle history narrations, as illustrated in Fig.~\ref{fig:context_len}(a), revealed:
\textbf{(i)} With the same number of clips, the \textcolor{NiceBlue}{$\blacktriangle$}vid+cap  variant surpasses \textcolor{Orange}{$\blacktriangle$}cap alone, whereas \textcolor{red}{$\blacktriangle$}vid exhibits negligible change. This emphasizes the significance of narrations in grasping event relationships and the challenge of reasoning in vision solely.
\textbf{(ii)} Increasing the number of input clips consistently leads to performance gains in the ‘vid+cap’ and ‘cap’ variants.

In ‘Video with Plots’ setting, our temporal \textcolor{darkgreen}{$\blacksquare$}estiamted strategy is contrasted with a variant that utilizes \textcolor{darkred}{$\blacksquare$}
clip vision-text similarities for filtering, as shown in Fig.~\ref{fig:context_len}(b). Our approach results highlighting the necessary of a reliable plot sampling method. Additionally, we observe a consistent trend of improved performance with increased context length, emphasizing the necessity of developing language models capable of processing longer contexts.

\begin{figure}
  \centering
  \begin{minipage}[t]{0.4\linewidth}
  \centering
\begin{tikzpicture}
    \begin{axis}[
        axis x line*=bottom,
        axis y line*=left,
        legend pos=north east,
        ymin=10, ymax=30,
        xmin=0, xmax=8,
        xticklabel={\pgfmathparse{\tick}\pgfmathprintnumber{\pgfmathresult}},
        xtick={0,2,4,6,8},
        ytick={10, 15, 20, 25, 30},
        ylabel={CIDER},
        yticklabel pos=left,
        ylabel style={font=\Huge, yshift=-15pt},
        xlabel style={font=\Huge},
        width=\linewidth,
        legend style={cells={align=left}},
        label style={font=\footnotesize},
        tick label style={font=\footnotesize},
        legend style={at={(0.03,1.0)},anchor=north west, font=\footnotesize}, 
    ]
    \addplot[mark=triangle*,style={thick},NiceBlue,mark size=3pt] plot coordinates {
        (0, 11.4)
        (2, 17.5)
        (4, 23.7)
        (6, 27.8)
        (8, 28.5)
    };
    \addlegendentry{vid+cap}

     \addplot[mark=triangle*,style={thick},orange,mark size=3pt] plot coordinates {
        (0, 11.4)
        (2, 13.1)
        (4, 15.4)
        (6, 18.4)
        (8, 18.9)
    };
    \addlegendentry{cap}

     \addplot[mark=triangle*,style={thick},red,mark size=3pt] plot coordinates {
        (0, 11.4)
        (2, 11.6)
        (4, 11.5)
        (6, 12.2)
        (8, 12.4)
    };
    \addlegendentry{vid}
    \end{axis}
\end{tikzpicture}
\captionsetup{labelformat=empty}
\vspace*{-1em}
\caption*{{(a) Diff. \# of history clip under diff. instruction template (iv). \vspace{-1em}}}
  \end{minipage}
  \hspace{1cm}
  \begin{minipage}[t]{0.4\linewidth}
  \centering
\begin{tikzpicture}
 	\begin{axis} [
		axis x line*=bottom,
		axis y line=left,
		legend pos=north east,
		ymin=10, ymax=20,
		xmin=0, xmax=512,
		xticklabel={\pgfmathparse{\tick}\pgfmathprintnumber{\pgfmathresult}},
		xtick={0,64,128,256,512},
		ytick={10, 13, 16, 19},
            ylabel={CIDER},
            yticklabel pos=right,
            ylabel style={font=\Huge, yshift=-15pt},
		xlabel style={font=\Huge},
		width=\linewidth,
		legend style={cells={align=left}},
		label style={font=\footnotesize},
		tick label style={font=\footnotesize},
        legend style={at={(0.03,1.0)},anchor=north west, font=\footnotesize}, 
		]

		\addplot[mark=square*,style={thick},darkgreen,mark size=3pt] plot coordinates {
        (0, 11.4)
        (64, 12.8)
        (128, 15.4)
        (256, 16.9)%
        (512, 17.6)
		};
      \addlegendentry{estimated}

		\addplot[mark=square*,style={thick},darkred,mark size=3pt] plot coordinates {
        (0, 11.4)
        (64, 11.3)
        (128, 11.7)
        (256, 12.9)%
        (512, 14.2)
		};
      \addlegendentry{clip sim.}

	\end{axis}
\end{tikzpicture}
\captionsetup{labelformat=empty}
\vspace*{-1em}
\caption*{{(b) Diff. input plot length (\# of tokens) under diff. strategy. \vspace{-1em}}}  
  \end{minipage}
  \captionsetup{font={small}}
  \caption{\small{\textbf{Effect of Data Construction on \mad~dataset.}}}
  \vspace{-2em}
    \label{fig:context_len}
\end{figure}

\subsection{Visualization.}
\begin{figure*}[!t]
    \centering
    \begin{subfigure}{.48\linewidth}
        \centering
        \includegraphics[width=\linewidth]{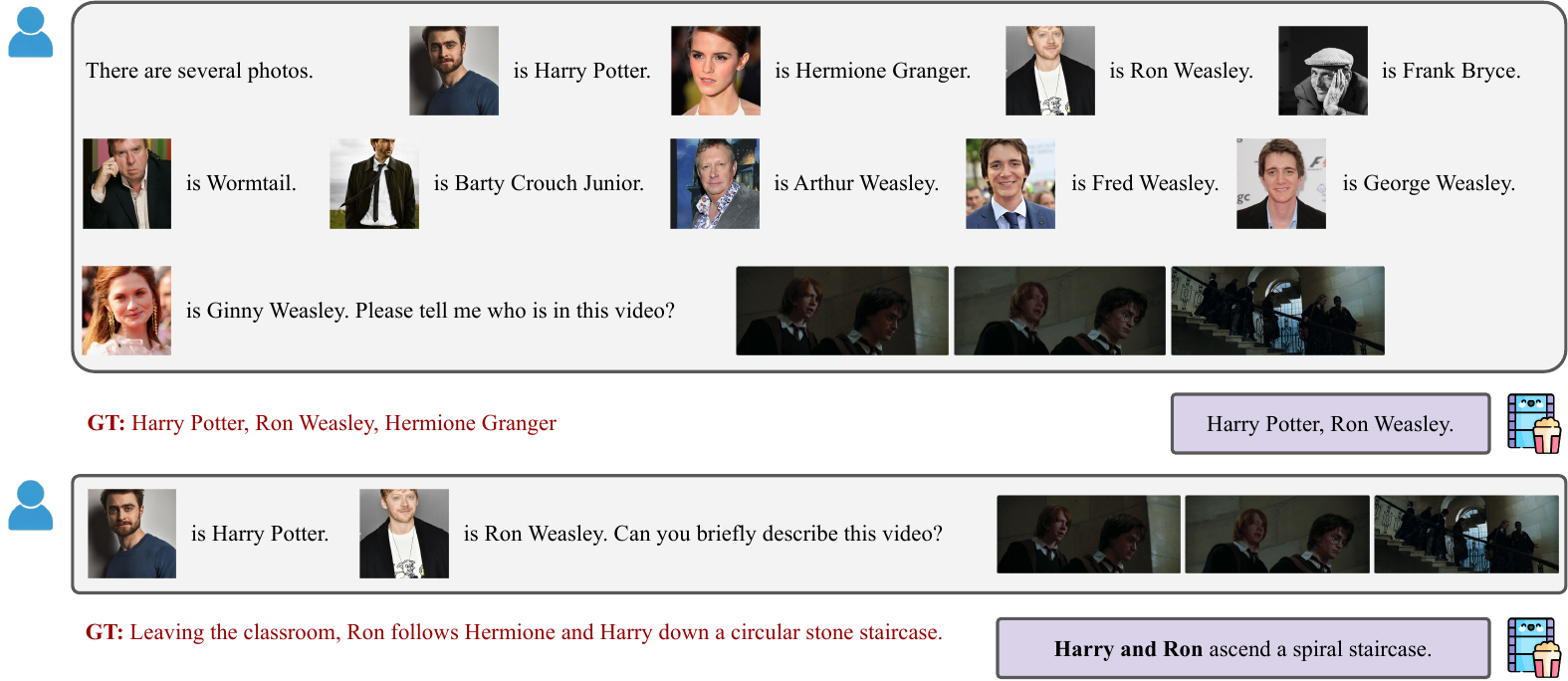}
        \caption{\textbf{Video-Images} instruction on MAD.} 
    \end{subfigure}
    \hfill 
    \begin{subfigure}{.48\linewidth}
        \centering
        \includegraphics[width=\linewidth]{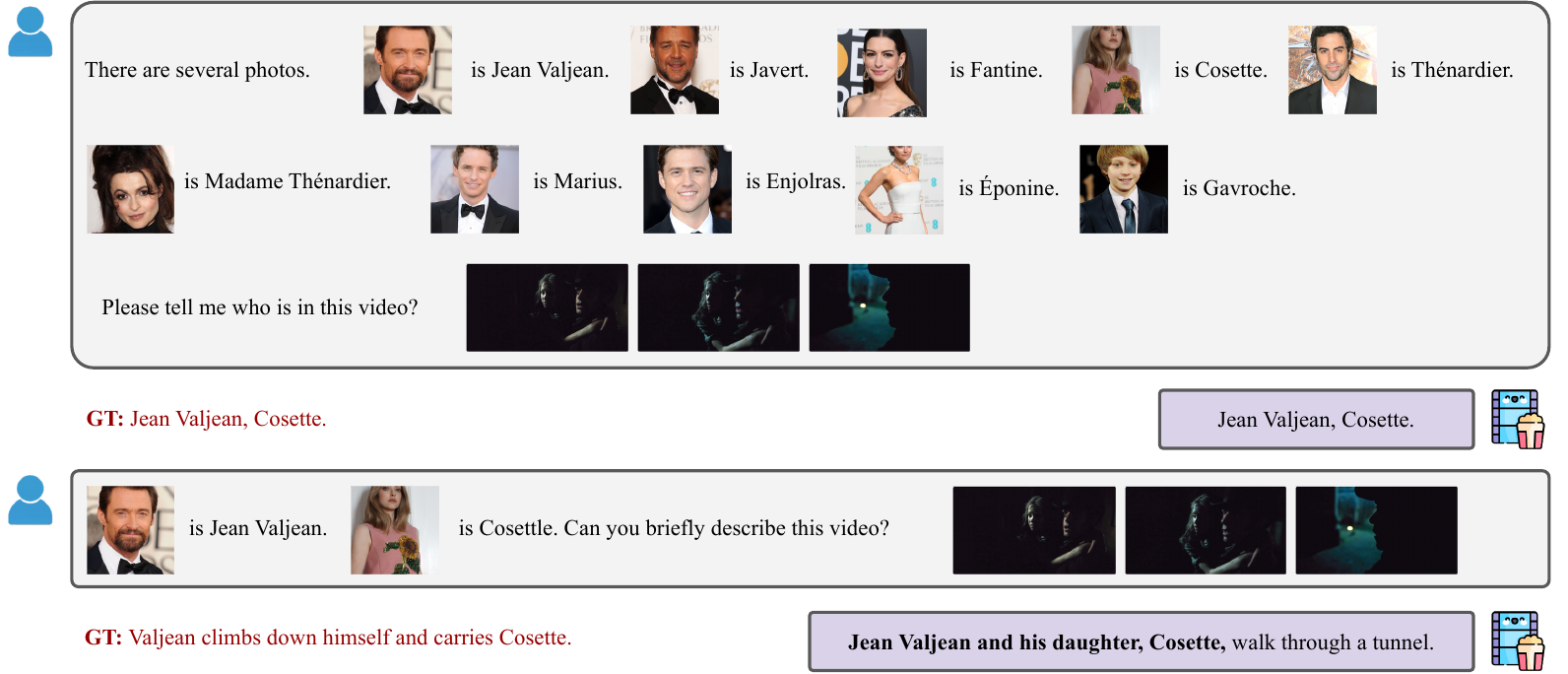}
        \caption{\textbf{Video-Images} instruction on MAD.} 
    \end{subfigure} 

    \begin{subfigure}{.48\linewidth}
        \centering
        \includegraphics[width=\linewidth]{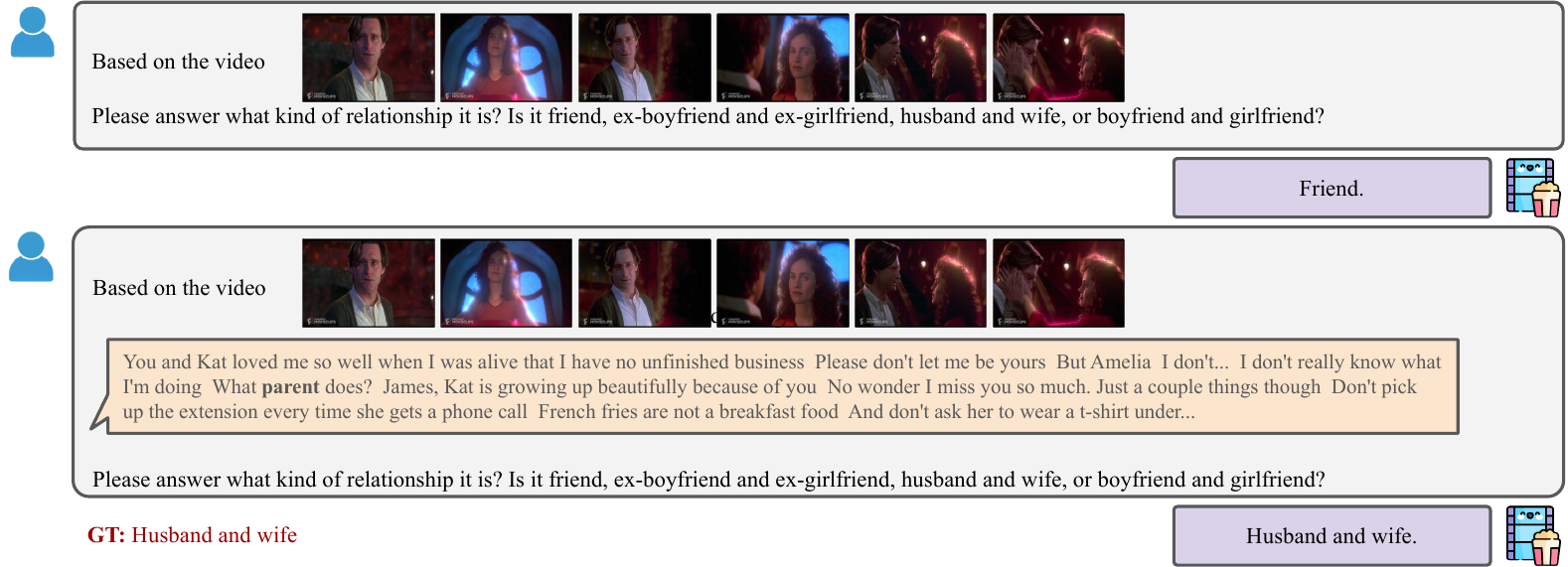}
        \caption{\textbf{Video-Subtitles} instruction on LVU. } 
    \end{subfigure}
    \hfill
    \begin{subfigure}{.48\linewidth}
        \centering
        \includegraphics[width=\linewidth]{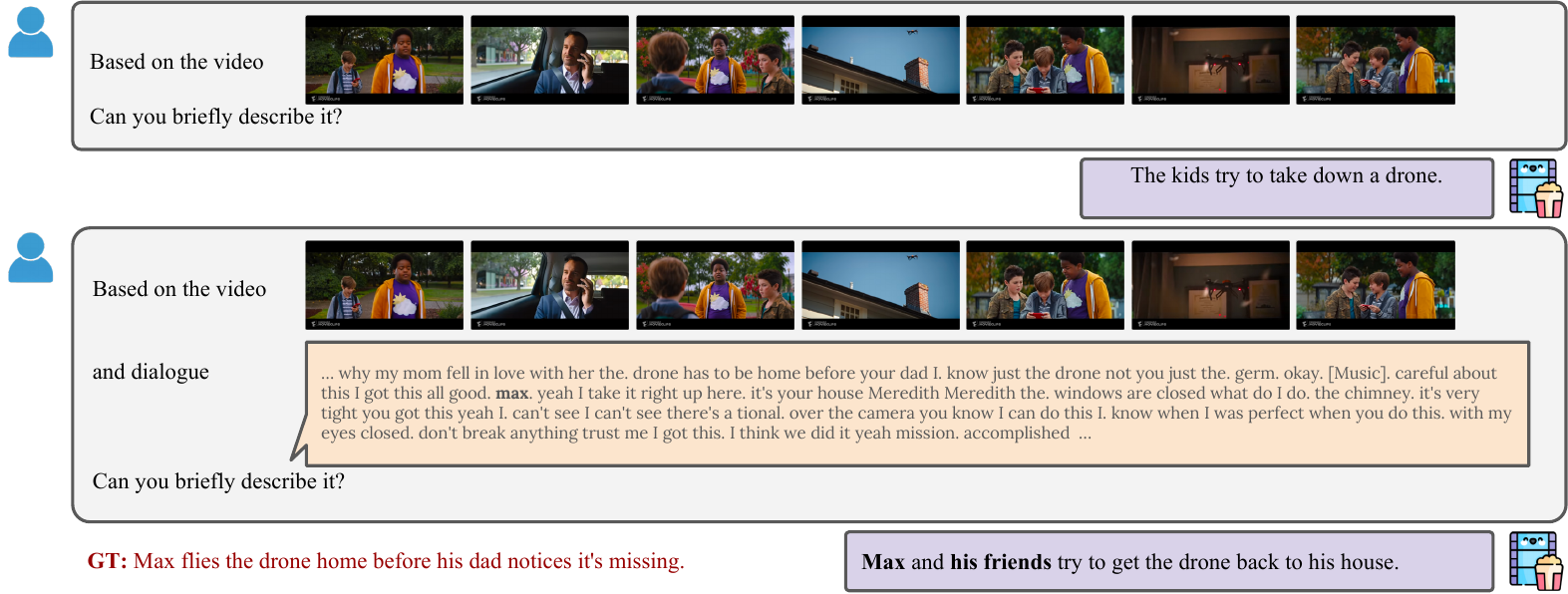}
        \caption{\textbf{Video-Subtitles} instruction on CMD.} 
    \end{subfigure}

    \begin{subfigure}{.48\linewidth}
        \centering
        \includegraphics[width=\linewidth]{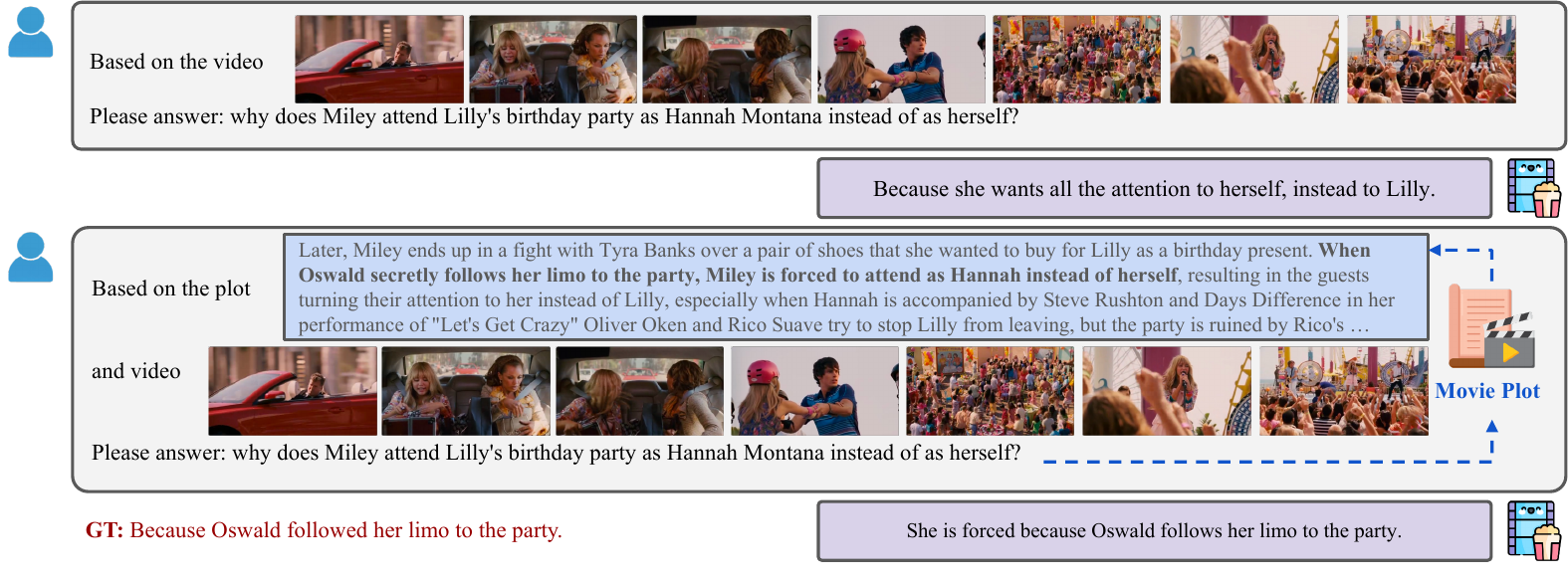}
        \caption{\textbf{Video-Plots} instruction on MovieQA.} 
    \end{subfigure}
    \hfill
    \begin{subfigure}{.48\linewidth}
        \centering
        \includegraphics[width=\linewidth]{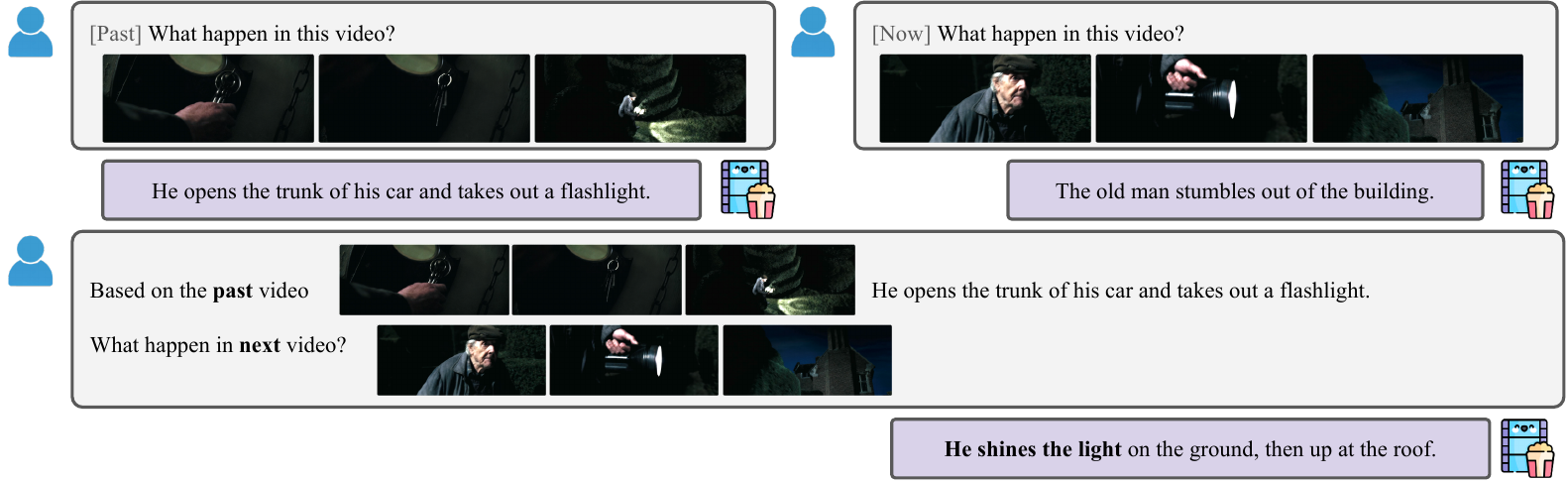}
        \caption{\textbf{Video-History} instruction on MAD.} 
    \end{subfigure}
\vspace{-.5em}
\caption{Visualization of \our~by providing different kinds of interleaved multimodal prompts for different applications.
\vspace{-2em}
}
\label{fig:vis}
\end{figure*}

In Fig.~\ref{fig:vis}, we provide visualization of \our~under different kinds of interleaved multimodal instruction. 
\textbf{(a-b) Video with Images.}
First, we provide the top-10 character photos and let the model decide who is present. Next, using these selected characters as a context, we can generate an accurate narration with character names. \our~can handle cases with multiple characters.
\textbf{(c-d) Video with Subtitles.}
We use subtitles to enhance the situational understanding. As seen in (c), it remains challenging to determine whether the relationship is that of friends or a couple from visual. Notably, as shown in (d), with the dialogue, the language model can associate the name (i.e., Max) with relationships (i.e., friends) and produce a meaningful caption.
\textbf{(e) Video with Plots:}
We sample an example from \movieqa. As shown, questions such as 'how' are primarily challenging to derive from the visual stream. Therefore, by using a retrieval-augmented plot as a reference (e.g., the background of why Miley attends the birthday party as Hannah Montana), the model can find the cue and derive the correct reason.
\textbf{(f) Video with History:}
In the multi-video settings, we use it to enhance the association of different events. As shown in (f), with the past prediction as context (e.g., “He opens the trunk of his car and takes out a flashlight”), the model can derive the next narration more naturally (e.g., “shines the light since he takes out a flashlight in the previous scene”).
\vspace{-0.4cm}
\section{Conclusion and Limitations}
\vspace{-0.3cm}
In this paper, we introduce {\our}, a video-language model designed for video context understanding. 
By modeling narrative videos as an interleaved multimodal sequence with external knowledge and additional modalities, \our~can support flexible interleaved multimodal instruction, and resolve several challenges including dialogues understanding, character identification, and event dependency. Furthermore, we illustrate how to collect the corresponding tuning data for each type of interleaved prompt. We demonstrate the effectiveness and flexibility of our \our~model across six diverse datasets. 

\noindent
Despite its strengths, our model still has \textit{limitations}.
We have not yet incorporated novel architectural designs, instead choosing to restrict visual branches (e.g., using CLS tokens instead of patch tokens) to save token length.
Exploring ways to effectively model long context without affecting visual input remains a topic worthy of discussion.
We leave them in our future work.

\noindent\textbf{Acknowledgements.}
This project is supported by the DSO National Laboratories. 
We extend our gratitude to Mattia Soldan for his assistance with MAD visualization examples, to Tengda Han's help with providing character identification modules and data, and to Makarand Tapaswi for his help on the MovieQA datasets.

\bibliographystyle{utils/splncs04}
\bibliography{main}

\end{document}